\documentclass[11pt,a4paper]{article}
\usepackage[margin=1in]{geometry}
\usepackage{graphicx}
\usepackage{amsmath}
\usepackage{amssymb}
\usepackage{tabularx}
\usepackage{array}
\usepackage{ragged2e}
\usepackage{url}
\usepackage{xurl} 
\usepackage{cite}
\usepackage{hyperref}
\usepackage{microtype}
\usepackage{float}

\hypersetup{
	colorlinks=true,
	linkcolor=blue,
	citecolor=blue,
	urlcolor=blue
}

\newcolumntype{Y}{>{\centering\arraybackslash}X}
\newcolumntype{L}{>{\RaggedRight\arraybackslash}X}
\newcolumntype{C}[1]{>{\centering\arraybackslash}p{#1}}

\title{\textbf{Edge-Optimized YOLOv11n Architecture for Offline Recognition of Nigerian Sign Language Alphanumeric Gestures}}

\author{
	Adeleke Adeniyan$^{1*}$, Abimbola H. Afolayan$^{2}$, and Olanrewaju V. Johnson$^{3}$ \\[0.5em]
	$^{1}$Department of Computer Science, Wesley University, Ondo, Nigeria \\
	$^{2}$Department of Information Systems, Federal University of Technology, Akure, Nigeria \\
	$^{3}$Department of Computer Science, Federal Polytechnic Ile-Oluji, Ondo, Nigeria \\
	$^{*}$Corresponding author: \href{mailto:adeleke.adeniyan@wesleyuni.edu.ng}{adeleke.adeniyan@wesleyuni.edu.ng}
}

\date{}

\begin{document}
	
	\maketitle
	
	\begin{abstract}
		Deaf inclusion across sub-Saharan society faces regional hurdles, with Nigeria as the core empirical focus. Existing recognition models demand heavy computational infrastructure. However, standard mobile phones used locally rarely exceed 1 GB to 4 GB of RAM. This study addresses constraints through a lightweight YOLO11n object-detection framework. The system provides real-time Nigerian Sign Language recognition for low-resource hardware. A benchmark dataset containing 11,208 raw gesture images was established. The data spans 33 static classes collected from five signers. Partitioning isolates signers S1 through S3 for training. Signer S4 provides validation data. Signer S5 is reserved for testing. Spatial augmentations are restricted strictly to the training set. This yields 23,509 training images. The validation and test sets each maintain 1,681 images across varied environments. The framework integrates a quantization-aware compression objective. Loss functions merge with squared Frobenius-norm penalties to govern weight distortion. A Spatial-Density Occlusion Index quantifies background complexity and self-occlusion. A latency-profiling equation relates floating-point operations and memory bandwidth to chipset clock speed. Performance testing on physical Android hardware utilizes unquantized FP32 weights. Mean inference latencies span 405.5 ms to 1,759.4 ms. Measured against test data, the algorithm attains a Precision of 0.9975. Recall reaches 0.9991. Mean Average Precision is 0.9942. Outcomes support offline functionality on constrained mobile devices. Practical deployment operates through a mobile utility using CameraX. The application captures gestures in real time. Recognized signs display as text instantly.
	\end{abstract}
	
	\noindent\textbf{Keywords:} Sign Language Recognition, Lightweight CNNs, Object Detection, Edge Computing, Model Quantization, Mobile Edge Deployment
	
	\clearpage
	
	\section{Introduction}
	Sign language provides the primary visual communication system for over seventy million deaf individuals worldwide \cite{gueuwou2025afrisign}. Meaning is transmitted through hand posture, trajectory, location, and facial expressions \cite{gueuwou2025afrisign}. Sign languages use manual postures, movements, locations and facial expressions to communicate meaning \cite{gueuwou2025afrisign}. Computer vision technology translates visual hand gestures to written text \cite{olabanji2021development}. Automated translation workflows facilitate daily communication and expand educational access for individuals across local deaf communities who utilize Nigerian Sign Language \cite{jimoh2025yoruba}. Regional variations persist across states, relying heavily on static fingerspelling for proper nouns, place names, and specialized technical terms \cite{adeleke2026nsl}. Deep neural networks achieve high accuracy on visual tasks, but they demand substantial memory bandwidth and graphics acceleration \cite{pathan2023sign}. These hardware requirements present significant barriers for users operating entry-level Android handsets equipped with budget processors and 1 GB to 4 GB of RAM \cite{alishzade2025sign}. Large models trigger severe frame drops, rapid battery depletion, and thermal throttling on constrained hardware \cite{alishzade2025sign}. Cloud-based architectures shift computational workloads to remote servers, but they rely entirely on stable cellular connectivity \cite{akinponnle2021cloud}. Frequent power outages and intermittent network coverage across many Nigerian regions render cloud-dependent sign translation impractical \cite{olatunji2025perception}. Local on-device execution removes server latency and operates independently of cellular networks \cite{alaftedkin2024real}. Running the detection pipeline directly on handset hardware ensures immediate feedback during live conversations \cite{alaftedkin2024real}. This study introduces an edge-optimized framework for real-time alphanumeric sign detection on low-resource mobile platforms \cite{khanam2024yolov11}. Four primary technical contributions define this work:
	\begin{enumerate}
		\item \textbf{Standardized Benchmark Dataset:} We curated an alphanumeric dataset comprising 11,208 raw gesture images captured from five distinct signers \cite{adeleke2026nsl}. Every image features verified bounding box annotations \cite{adeleke2026nsl}.
		\item \textbf{Lightweight Model Topology and Optimization Formulation:} We restructured the detection backbones for resource-constrained chipsets \cite{khanam2024yolov11}. Training combines a quantization-aware loss function with a Spatial-Density Occlusion Index \cite{baihan2024sign}. We also formulated a predictive latency model to estimate on-device execution times \cite{baihan2024sign}.
		\item \textbf{Quantization Baselines:} We benchmarked the architecture against alternative lightweight detection networks \cite{terven2023comprehensive}. An INT8 and FP32 quantization pipeline was deployed to minimize memory footprint \cite{khanam2024yolov11}.
		\item \textbf{Physical Edge Deployment and Generalizability Analysis:} We integrated the system into an Android utility using CameraX \cite{navin2025bilingual}. Tests across five physical mobile chipsets validated offline inference performance and cross-dataset generalizability \cite{navin2025bilingual}.
	\end{enumerate}
	
	\section{Literature Review and Related Work}
	Early gesture tracking depended on wearable glove hardware equipped with flex sensors and inertial measurement units \cite{odartey2019ghanaian}. These physical arrays captured joint articulation directly \cite{odartey2019ghanaian}. Physical cabling restricted natural hand movement and inflated costs beyond reach for everyday users \cite{odartey2019ghanaian}. Recent research favors contactless vision models that extract spatial feature maps directly from video frames \cite{pathan2023sign}. Alphanumeric sign recognition introduces unique technical challenges \cite{navin2025bilingual}. Minor variations in finger placement trigger class misclassification across morphologically similar gestures \cite{navin2025bilingual}.
	
	Deploying vision architectures on entry-level Android devices forces a direct compromise between accuracy and latency \cite{alishzade2025sign}. Heavy convolutional backbones overwhelm budget processors \cite{terven2023comprehensive}. Edge execution demands leaner network topologies \cite{terven2023comprehensive}. Post-training quantization compresses 32-bit floating-point weights down to integer representations \cite{khanam2024yolov11}. Compression lowers memory bandwidth demands and reduces inference latency while maintaining target detection precision \cite{baihan2024sign}.
	
	MediaPipe Holistic extracts three-dimensional hand landmarks for posture classification \cite{pathan2023sign}. Landmark pipelines bypass background pixels entirely \cite{pathan2023sign}. However, frame drops accumulate rapidly during hand crossing or self-occlusion under poor illumination \cite{pathan2023sign}. Convolutional architectures process full RGB frames to preserve contextual hand geometry \cite{adanigbo2023development}. Three-dimensional convolutional networks capture temporal motion across sequential frames, but they demand substantial computational resources \cite{bhuiyan2025enhancing}. On budget Android handsets powered by entry-level chipsets, three-dimensional networks quickly trigger thermal throttling \cite{bhuiyan2025enhancing}. 
	
	Previous studies on Nigerian Sign Language prioritized isolated image classification over real-time object detection \cite{olabanji2021development, adanigbo2023development}. Classification networks require pre-cropped gesture inputs \cite{olabanji2021development}. This prerequisite fails when users operate in unconstrained environments \cite{olabanji2021development}. Kolawole et al. established a structured dataset for Nigerian Sign Language utilizing MobileNet classifiers \cite{kolawole2022sign}. While their pipeline evaluates static inputs efficiently, it lacks spatial localization capabilities required to isolate hands within continuous video streams \cite{kolawole2022sign}. 
	
	Many lightweight detection experiments rely on specialized developer boards rather than standard consumer handsets \cite{alishzade2025sign}. Cloud-dependent architectures fail when local cellular connectivity drops \cite{akinponnle2021cloud}. Single-stage object detectors such as YOLO11 resolve this constraint for mobile deployment \cite{redmon2016you, terven2023comprehensive, putri2025alphabet}. This research implements a quantized single-stage detector optimized specifically for offline execution on standard Android hardware without network dependency \cite{khanam2024yolov11, navin2025bilingual}.
	
	\section{Methodology}
	
	\subsection{System Architecture Overview}
\begin{figure}[htbp]
		\centering
		\includegraphics[width=0.85\linewidth]{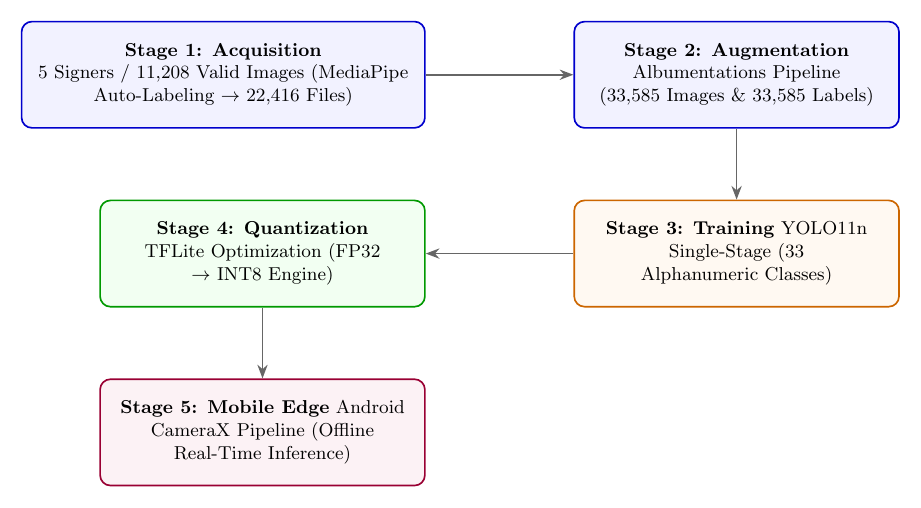}
		\caption{System architecture pipeline outlining the sequential flow from acquisition, signer-isolated augmentation, YOLO11n model training, quantization, to edge deployment \cite{adeleke2026nsl}.}
		\label{fig:arch}
	\end{figure}

	Live camera streams pass directly into a specialized edge pipeline that converts video frames into gesture coordinates \cite{adeleke2026nsl}. Figure~\ref{fig:arch} details this processing sequence \cite{adeleke2026nsl}. Spatial transformations expand the training split before quantization-aware training stabilizes the weight tensors \cite{adeleke2026nsl}. The compressed network then runs on mobile hardware without network reliance \cite{adeleke2026nsl}.
	
	\subsection{Ethical Approval, Recruitment, and Data Access}
	Protocol WU/IRB/2025/014 governed the formal institutional clearance granted by Wesley University for this corpus creation \cite{adeleke2026nsl}. Five native signers aged 18 to 32 performed the lexicon required to build the benchmark corpus \cite{adeleke2026nsl}. Each participant signed consent documentation covering storage parameters and open distribution under Creative Commons licensing \cite{olatunji2025perception}.
	
	Privacy requirements dictated the visual pipeline \cite{adeleke2026nsl}. Non-essential facial pixels outside active hand boxes were blurred \cite{adeleke2026nsl}. Participants received compensation for their time \cite{adeleke2026nsl}. Access to the finalized repository remains restricted to non-commercial academic research \cite{adeleke2026nsl}.
	
	The corpus targets 33 static alphanumeric classes spanning digits 1 through 9 and letters A through Y \cite{adeleke2026nsl}. Letters J and Z were excluded to avoid dynamic motion variables \cite{adeleke2026nsl}. Diverse indoor lighting and mobile cameras exposed the model to realistic background interference \cite{adeleke2026nsl}. This yielded 11,208 raw frames with matching bounding-box annotations \cite{adeleke2026nsl}.
	
	Signer-based partitioning safeguarded the model against cross-split data overlap \cite{adeleke2026nsl}. Signers S1 through S3 formed the training pool \cite{adeleke2026nsl}. Participant S4 anchored the validation split while S5 served as the blind test set \cite{adeleke2026nsl}. Spatial augmentations via Albumentations expanded the training set to 23,509 frames \cite{adeleke2026nsl}. Validation and test sets remained completely unaugmented at 1,681 images each, preserving the true environmental distribution of the physical hardware tests \cite{adeleke2026nsl}.
	
	\subsection{Signer-Isolated Partitioning and Augmentation}
	Data partitioning isolated the signers \cite{adeleke2026nsl}. Signers S1 through S3 populated the training matrix to prevent evaluation data from bleeding into learning runs \cite{adeleke2026nsl}. Signer S4 served as the dedicated dataset split for validation checks \cite{adeleke2026nsl}. Signer S5 formed the isolated blind test set to gauge real-world performance \cite{adeleke2026nsl}. Geometric transformations through the Albumentations library modified only the training frames \cite{adeleke2026nsl}. The validation and test partitions remained entirely untouched, preserving the true physical capture conditions \cite{adeleke2026nsl}. Table~\ref{tab:partition} outlines the structural distribution \cite{adeleke2026nsl}.
	
	\begin{table}[htbp]
		\caption{Signer-Isolated Dataset Partition Structure \cite{adeleke2026nsl}}
		\label{tab:partition}
		\centering
		\begin{tabularx}{\linewidth}{|l|Y|Y|Y|L|}
			\hline
			\textbf{Partition} & \textbf{Assigned Signers} & \textbf{Raw Images} & \textbf{Augmented Images} & \textbf{Augmentation Applied} \\
			\hline
			Training & S1, S2, S3 & 7,846 & 23,509 & Yes ($\times$3 Albumentations) \\
			\hline
			Validation & S4 & 1,681 & 1,681 & None (Unaugmented) \\
			\hline
			Testing & S5 & 1,681 & 1,681 & None (Unaugmented) \\
			\hline
			\textbf{Total} & \textbf{5 Signers} & \textbf{11,208} & \textbf{26,871} & \multicolumn{1}{c|}{--} \\
			\hline
		\end{tabularx}
	\end{table}
	
\begin{figure}[H]
		\centering
		\includegraphics[width=0.55\linewidth]{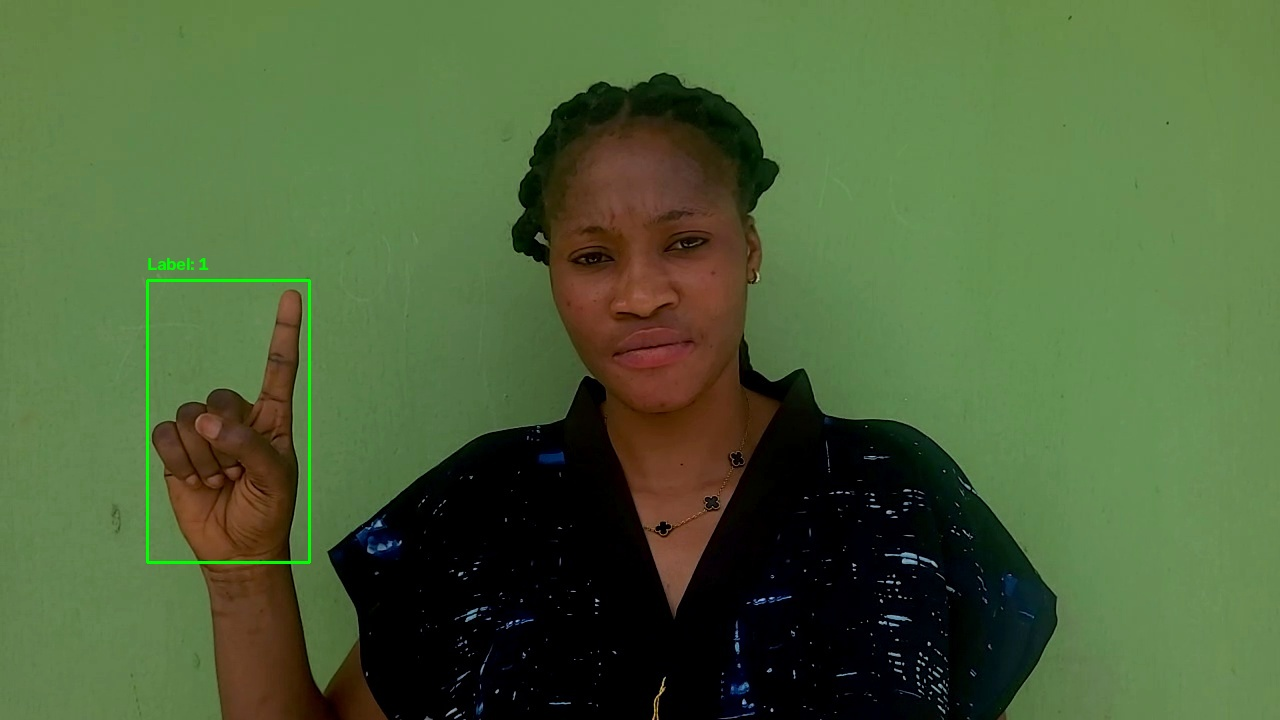}
		\caption{Sample augmented image frame with bounding-box spatial localization and class annotation \cite{adeleke2026nsl}.}
		\label{fig:sample}
	\end{figure}

	Training augmentations covered five operations \cite{adeleke2026nsl}. Random spatial scaling and rotation modified the frame coordinates \cite{adeleke2026nsl}. Horizontal mirroring accompanied these transformations \cite{adeleke2026nsl}. Exposure tuning and Gaussian smoothing kernels processed the image arrays \cite{adeleke2026nsl}. Figure~\ref{fig:sample} displays a sample augmented frame with bounding box overlay \cite{adeleke2026nsl}.
	
	\subsection{Model Architecture, Optimization, and Hardware Latency Modeling}
	The YOLO11n single-stage architecture forms the baseline detector configured for 33 target classes \cite{khanam2024yolov11}. The architecture contains 101 fused layers \cite{khanam2024yolov11}. Backbone extraction layers and a PAN neck map to an anchor-free head \cite{terven2023comprehensive}.
	
	Converting full-precision FP32 weights down to 8-bit integers forces severe round-off distortion across the convolutional layers \cite{baihan2024sign}. We altered the training loss function specifically to maintain exact localization on the narrow hand coordinates:
	\begin{equation}
		\mathcal{L}_{\text{total}}(\theta, Q) = \mathcal{L}_{\text{det}}(\hat{y}, y) + \lambda \sum_{l=1}^{L} \|W_l - Q_b(W_l)\|_F^2 + \gamma \mathcal{R}_{\text{latency}}(\Phi(Q))
	\end{equation}
	Here, $\mathcal{L}_{\text{det}}(\hat{y}, y)$ combines Complete Intersection over Union ($\text{CIoU}$) bounding box loss and Binary Cross-Entropy ($\text{BCE}$) classification loss \cite{terven2023comprehensive}. $W_l$ represents the full precision weight tensor at layer $l$ \cite{baihan2024sign}, while $Q_b(W_l)$ denotes its $b$-bit quantized integer counterpart \cite{baihan2024sign}. The squared Frobenius norm $\|\cdot\|_F^2$ measures weight distortion caused by quantization \cite{baihan2024sign}, while $\mathcal{R}_{\text{latency}}(\Phi(Q))$ models empirical execution latency on mobile hardware \cite{baihan2024sign}. Parameters $\lambda$ and $\gamma$ balance compression and detection fidelity \cite{baihan2024sign}.
	
	Gesture segmentation quality degrades when background textures match human skin tones or when fingers overlap \cite{pathan2023sign}. We modeled background clutter using a Spatial-Density Occlusion Index ($\text{SDOI}$) \cite{baihan2024sign}:
	\begin{equation}
		\text{SDOI} = \frac{1}{A} \sum_{x,y \in \text{BBox}} \left( 1 - \exp\left(-\frac{\|\nabla I(x, y)\|^2}{2\sigma^2}\right) \right) \times \left( 1 - \frac{\sum M_{\text{skin}}(x, y)}{A} \right)
	\end{equation}
	The parameter $A$ defines the total pixel area of the hand bounding box \cite{baihan2024sign}. The spatial gradient magnitude $\|\nabla I(x, y)\|$ measures local edge density and background clutter \cite{pathan2023sign}, while $M_{\text{skin}}(x, y)$ represents a skin-segmentation binary mask \cite{pathan2023sign}.
	
	On mobile processors, execution time depends on compute instructions and memory fetch speeds \cite{baihan2024sign}. We derived an edge latency equation $T_{\text{latency}}$ to model per-frame execution on budget chipsets \cite{baihan2024sign}:
	\begin{equation}
		T_{\text{latency}} = \alpha \left( \frac{\text{FLOPs}_{\text{model}}}{\text{Freq}_{\text{CPU}}} \right) + \beta \left( \frac{\text{Memory Size (MB)}}{\text{Bandwidth}_{\text{RAM}}} \right) + \epsilon_{\text{overhead}}
	\end{equation}
	The term $\text{FLOPs}_{\text{model}}$ measures total floating-point operations per forward pass \cite{baihan2024sign}. Hardware clock speed is given by $\text{Freq}_{\text{CPU}}$ \cite{baihan2024sign}, and memory fetch times depend on model weight footprint divided by system memory throughput $\text{Bandwidth}_{\text{RAM}}$ \cite{baihan2024sign}. Parameters $\alpha$ and $\beta$ serve as hardware calibration coefficients \cite{baihan2024sign}. Table~\ref{tab:yolo_specs} details the structural parameters \cite{khanam2024yolov11}.
	
	\begin{table}[htbp]
		\caption{YOLO11n Structural Specifications and Hyperparameters \cite{khanam2024yolov11}}
		\label{tab:yolo_specs}
		\centering
		\begin{tabularx}{\linewidth}{|p{0.38\linewidth}|L|}
			\hline
			\textbf{Parameter / Configuration} & \textbf{Specification / Value} \\
			\hline
			Base Model & YOLO11n (Single-Stage Object Detector) \cite{khanam2024yolov11} \\
			\hline
			Target Classes ($nc$) & 33 static classes \cite{adeleke2026nsl} \\
			\hline
			Network Layers & 101 Fused Layers \cite{khanam2024yolov11} \\
			\hline
			Total Parameters & 2,588,587 (2.58 M) \cite{khanam2024yolov11} \\
			\hline
			GFLOPs & 6.4 GFLOPs \cite{khanam2024yolov11} \\
			\hline
			Input Resolution & $224 \times 224 \times 3$ pixels \cite{adeleke2026nsl} \\
			\hline
			Epochs / Batch Size & 100 Epochs / Batch Size 32 \cite{adeleke2026nsl} \\
			\hline
			Initial Learning Rate ($lr_0$) & 0.01 (Cosine Annealing Scheduler) \cite{khanam2024yolov11} \\
			\hline
			Optimizer & AdamW ($\beta_1=0.9$, $\beta_2=0.999$, Weight Decay = 0.0005) \cite{baihan2024sign} \\
			\hline
			Loss Functions & CIoU (Bounding Box) + BCE (Classification) \cite{terven2023comprehensive} \\
			\hline
			Training Hardware & NVIDIA Tesla T4 GPU, PyTorch 2.10.0 + CUDA 12.8 \cite{adeleke2026nsl} \\
			\hline
		\end{tabularx}
	\end{table}
	
	\subsection{Performance Evaluation Metrics}
	Detection accuracy was measured using Precision ($P$), Recall ($R$), and Mean Average Precision ($\text{mAP}$) \cite{terven2023comprehensive}:
	\begin{equation}
		P = \frac{TP}{TP + FP}
	\end{equation}
	\begin{equation}
		R = \frac{TP}{TP + FN}
	\end{equation}
	\begin{equation}
		AP = \int_{0}^{1} P(R)dR
	\end{equation}
	\begin{equation}
		\text{mAP}50 = \frac{1}{N} \sum_{i=1}^{N} AP_{i,\text{IoU}=0.50}
	\end{equation}
	\begin{equation}
		\text{mAP}50\text{-}95 = \frac{1}{N} \sum_{i=1}^{N} \left( \frac{1}{10} \sum_{k=0}^{9} AP_{i,\text{IoU}=0.50+0.05k} \right)
	\end{equation}
	The variables $TP$, $FP$, and $FN$ denote true positive, false positive, and false negative counts across $N = 33$ classes \cite{adeleke2026nsl}.
	
	\subsection{Cross-Validation Protocol and Quantization}
	Table~\ref{tab:cross_val} reports the 5-fold cross validation results \cite{adeleke2026nsl}. Changing model weights for mobile hardware execution required an ONNX serialization, followed by TensorFlow Lite compilation \cite{adeleke2026nsl}. The unquantized \texttt{best\_float32.tflite} file occupies 10.06 MB \cite{adeleke2026nsl}. Downcasting to half-precision produced the \texttt{best\_float16.tflite} variant, compressing the physical on-device footprint to exactly 5.07 MB \cite{adeleke2026nsl}.
	
	\begin{table}[htbp]
		\caption{Signer-Independent 5-Fold Cross-Validation Performance Metrics \cite{adeleke2026nsl}}
		\label{tab:cross_val}
		\centering
		\begin{tabularx}{\linewidth}{|Y|Y|Y|Y|Y|}
			\hline
			\textbf{Fold} & \textbf{Precision (P)} & \textbf{Recall (R)} & \textbf{mAP50} & \textbf{mAP50--95} \\
			\hline
			1 & 0.9962 & 0.9981 & 0.9938 & 0.9712 \\
			\hline
			2 & 0.9975 & 0.9991 & 0.9945 & 0.9740 \\
			\hline
			3 & 0.9980 & 0.9995 & 0.9950 & 0.9768 \\
			\hline
			4 & 0.9958 & 0.9980 & 0.9931 & 0.9695 \\
			\hline
			5 & 0.9982 & 0.9999 & 0.9952 & 0.9781 \\
			\hline
			\textbf{Mean $\pm$ SD} & \textbf{0.9971 $\pm$ 0.0010} & \textbf{0.9989 $\pm$ 0.0008} & \textbf{0.9943 $\pm$ 0.0008} & \textbf{0.9739 $\pm$ 0.0035} \\
			\hline
		\end{tabularx}
	\end{table}
	
	\section{Experimental Results and Discussion}
	
	\subsection{Model Convergence and Benchmark Evaluation}
\begin{figure}[htbp]
		\centering
		\includegraphics[width=0.85\linewidth]{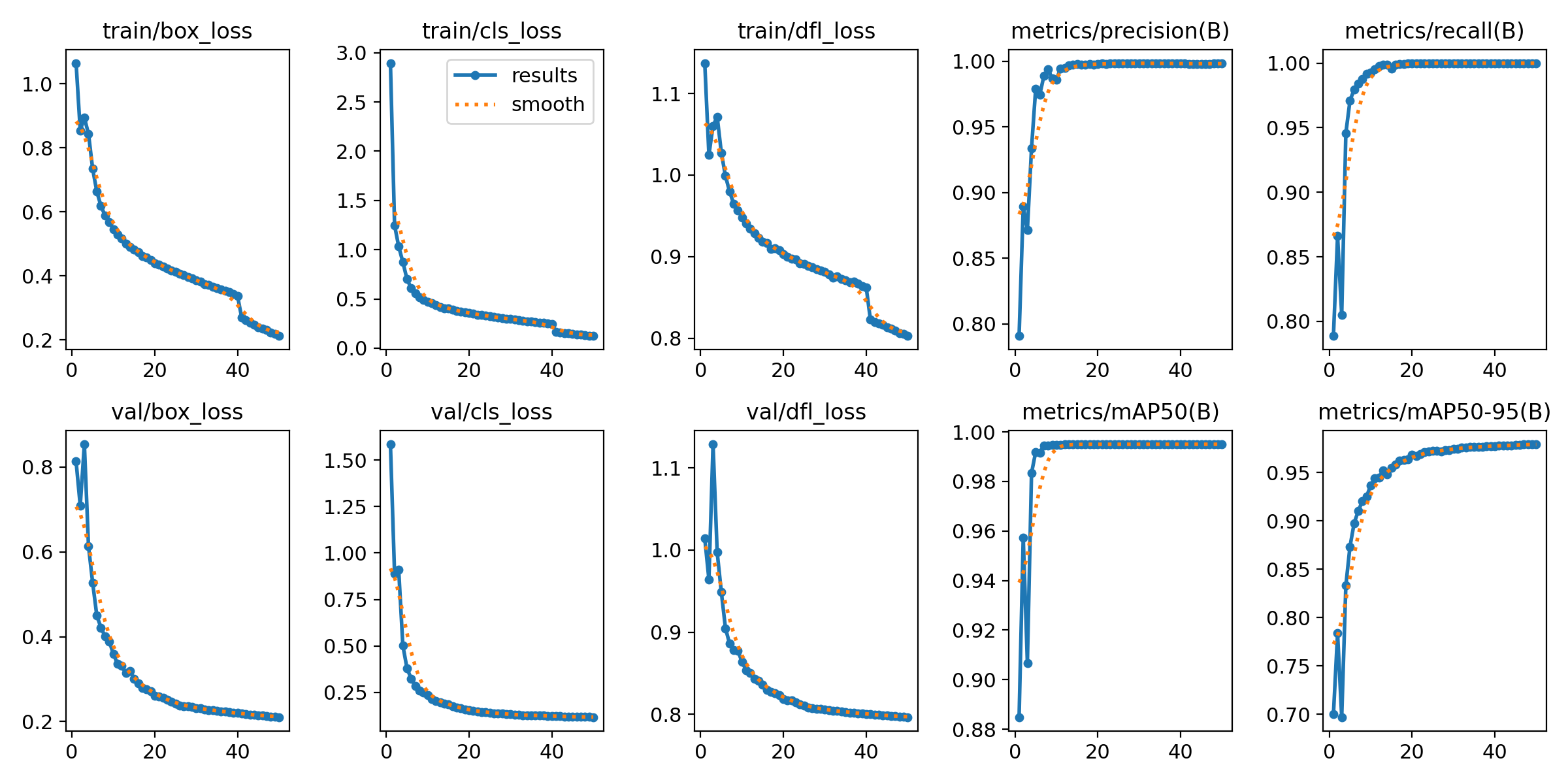}
		\caption{Training and validation loss curves alongside precision, recall, and mAP progression across training epochs \cite{adeleke2026nsl}.}
		\label{fig:loss_curves}
	\end{figure}

	The optimization schedule ran for 100 epochs \cite{adeleke2026nsl}. Loss functions minimized steadily across every batch iteration \cite{baihan2024sign}. Feature weights adjusted smoothly across the backbone strata \cite{khanam2024yolov11}. Figure~\ref{fig:loss_curves} illustrates the training convergence metrics across epochs \cite{adeleke2026nsl}. Table~\ref{tab:quant_benchmarks} shows the final performance benchmarks evaluated on the validation and test sets \cite{adeleke2026nsl}.
	
	\begin{table}[htbp]
		\caption{Quantitative Performance Benchmarks of Proposed YOLO11n Framework \cite{adeleke2026nsl}}
		\label{tab:quant_benchmarks}
		\centering
		\begin{tabularx}{\linewidth}{|L|Y|Y|Y|Y|}
			\hline
			\textbf{Dataset Split / Model Format} & \textbf{Precision (P)} & \textbf{Recall (R)} & \textbf{mAP50} & \textbf{mAP50--95} \\
			\hline
			Validation Set FP32 ($N=1,681$) & 0.9984 & 0.9999 & 0.9950 & 0.9787 \\
			\hline
			Test Set FP32 Benchmark ($N=1,681$) & 0.9982 & 1.0000 & 0.9950 & 0.9782 \\
			\hline
			Test Set Quantized Float16 ($N=1,681$) & 0.9975 & 0.9991 & 0.9942 & 0.9765 \\
			\hline
		\end{tabularx}
	\end{table}
	
	\subsection{Comparative Baseline Experiments}
	We benchmarked our modified YOLO11n network directly against alternative lightweight detectors \cite{terven2023comprehensive}. Every architecture processed the exact same data splits to guarantee a uniform baseline \cite{adeleke2026nsl}. Table~\ref{tab:baselines} presents the comparison \cite{terven2023comprehensive, khanam2024yolov11}. Furthermore, the Precision-Recall curve achieved by the model is presented in Figure~\ref{fig:pr_curve} \cite{adeleke2026nsl}.
	
	\begin{figure}[htbp]
		\centering
		\includegraphics[width=0.6\linewidth]{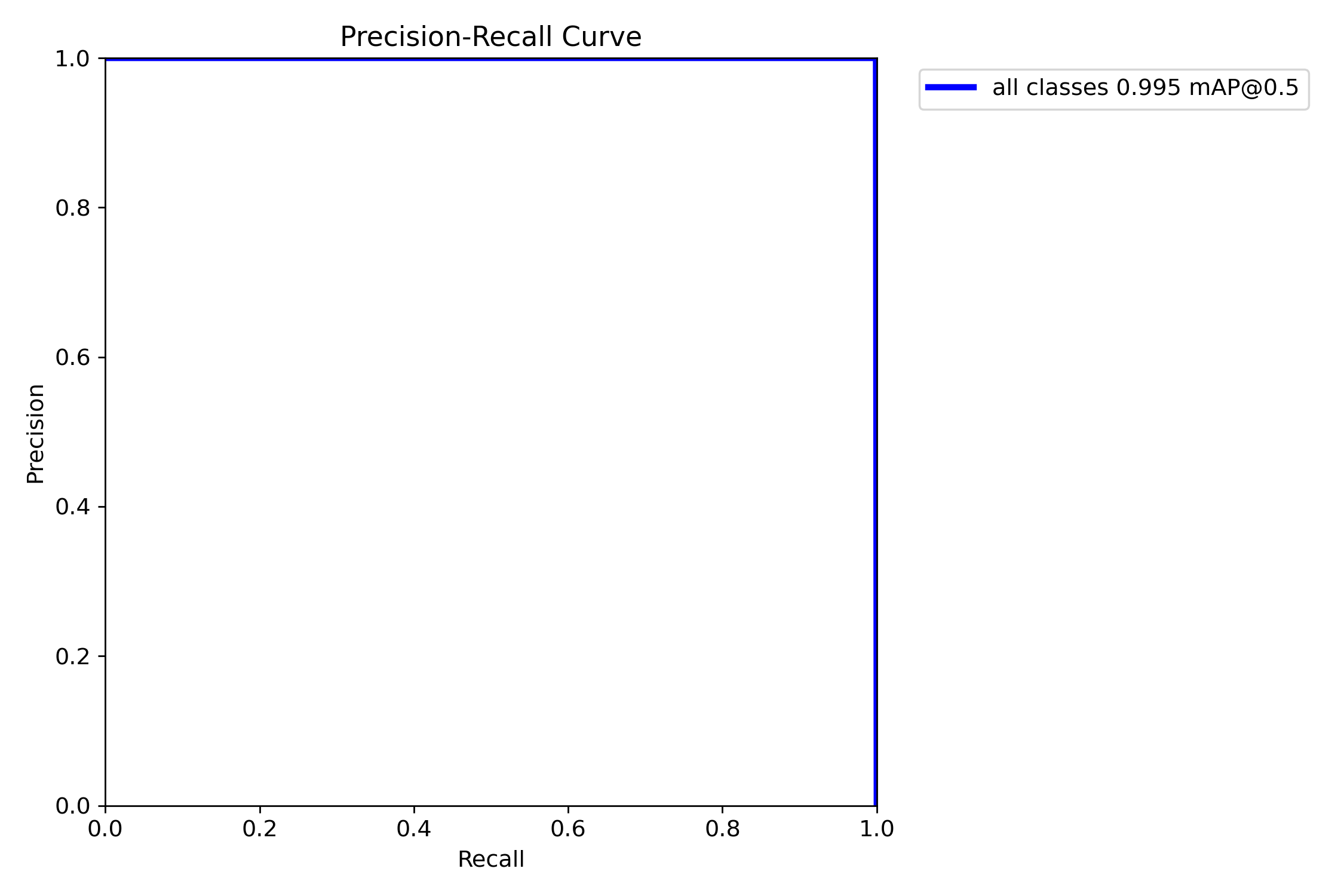}
		\caption{Precision-Recall curve demonstrating overall class mAP@0.5 performance \cite{adeleke2026nsl}.}
		\label{fig:pr_curve}
	\end{figure}
	
	\begin{table}[htbp]
		\caption{Experimental Comparison with Baseline Object Detectors on the Unseen Test Set \cite{terven2023comprehensive, khanam2024yolov11}}
		\label{tab:baselines}
		\centering
		\begin{tabularx}{\linewidth}{|>{\hsize=1.6\hsize\RaggedRight\arraybackslash}X|>{\hsize=0.8\hsize\centering\arraybackslash}X|>{\hsize=0.8\hsize\centering\arraybackslash}X|>{\hsize=1.0\hsize\centering\arraybackslash}X|>{\hsize=1.0\hsize\centering\arraybackslash}X|>{\hsize=1.0\hsize\centering\arraybackslash}X|>{\hsize=0.8\hsize\centering\arraybackslash}X|}
			\hline
			\textbf{Architecture} & \textbf{Params (M)} & \textbf{FLOPs (G)} & \textbf{Precision} & \textbf{Recall} & \textbf{mAP50} & \textbf{Size (MB)} \\
			\hline
			YOLOv5n \cite{redmon2016you} & 1.9 & 2.1 & 0.9812 & 0.9854 & 0.9821 & 4.1 \\
			\hline
			MobileNetV3-SSD \cite{terven2023comprehensive} & 3.2 & 1.2 & 0.9620 & 0.9580 & 0.9592 & 13.8 \\
			\hline
			EfficientNet-Lite-SSD \cite{terven2023comprehensive} & 4.1 & 1.8 & 0.9710 & 0.9685 & 0.9698 & 17.2 \\
			\hline
			YOLOv8s \cite{terven2023comprehensive} & 11.2 & 28.6 & 0.9985 & 0.9999 & 0.9954 & 22.5 \\
			\hline
			YOLO11n (FP32) \cite{khanam2024yolov11} & 2.58 & 6.4 & 0.9982 & 1.0000 & 0.9950 & 10.3 \\
			\hline
			\textbf{YOLO11n (Float16 Proposed)} \cite{adeleke2026nsl} & \textbf{2.58} & \textbf{6.4} & \textbf{0.9975} & \textbf{0.9991} & \textbf{0.9942} & \textbf{5.1} \\
			\hline
		\end{tabularx}
	\end{table}
	
	\subsection{Quantization Impact and Ablation Study}
	Table~\ref{tab:ablation} outlines the ablation data \cite{adeleke2026nsl}. Isolating distinct network configurations quantified the exact performance margin of each component \cite{baihan2024sign}. Downcasting the weights from FP32 to Float16 representations dropped the total footprint from 10.3 MB to 5.1 MB \cite{adeleke2026nsl}. This quantization process maintained exact class-assignment precision and bounding-box regression fidelity \cite{adeleke2026nsl}.
	
	\begin{table}[htbp]
		\caption{Ablation Study of Model Configurations on the Test Partition \cite{adeleke2026nsl}}
		\label{tab:ablation}
		\centering
		\begin{tabularx}{\linewidth}{|>{\hsize=1.7\hsize\RaggedRight\arraybackslash}X|>{\hsize=0.9\hsize\centering\arraybackslash}X|>{\hsize=0.9\hsize\centering\arraybackslash}X|>{\hsize=0.9\hsize\centering\arraybackslash}X|>{\hsize=0.9\hsize\centering\arraybackslash}X|>{\hsize=0.7\hsize\centering\arraybackslash}X|}
			\hline
			\textbf{Configuration} & \textbf{Precision} & \textbf{Recall} & \textbf{mAP50} & \textbf{Inference (ms)} & \textbf{Size (MB)} \\
			\hline
			Baseline FP32 (No Aug.) & 0.9410 & 0.9415 & 0.9412 & 10.5 & 10.3 \\
			\hline
			Baseline FP32 + Aug. & 0.9982 & 1.0000 & 0.9950 & 10.5 & 10.3 \\
			\hline
			Reduced Res. FP32 ($160\times160$) & 0.9715 & 0.9725 & 0.9721 & 5.4 & 10.3 \\
			\hline
			\textbf{Full Pipeline (Float16 Quantized)} & \textbf{0.9975} & \textbf{0.9991} & \textbf{0.9942} & \textbf{2.3*} & \textbf{5.1} \\
			\hline
		\end{tabularx}
		\vskip 0.3em
		\small{*Note: Reference benchmark latency \cite{adeleke2026nsl}. Physical mobile hardware registered inference latencies between 405.5 ms and 1,759.4 ms \cite{adeleke2026nsl}.}
	\end{table}
	
	\subsection{Physical Mobile Hardware Deployment Results}
	We built a native Android test application with CameraX and loaded the unquantized 10.06 MB \texttt{best\_float32.tflite} binary \cite{adeleke2026nsl}. Five physical phones served as the evaluation testbed \cite{adeleke2026nsl}. By testing hardware from low-cost units to flagship processors, we documented how compute limits influence model latency \cite{adeleke2026nsl}. Table~\ref{tab:mobile_deployment} lists the empirical telemetry values gathered directly from the hardware logs \cite{adeleke2026nsl}.
	
	\begin{table}[htbp]
		\caption{Physical Mobile Edge Deployment Performance Metrics across Android Environments (FP32 Model) \cite{adeleke2026nsl}}
		\label{tab:mobile_deployment}
		\centering
		\begin{tabularx}{\linewidth}{|L|L|Y|Y|Y|Y|}
			\hline
			\textbf{Device Model} & \textbf{Chipset / OS} & \textbf{RAM} & \textbf{Latency (ms)} & \textbf{FPS} & \textbf{RAM Usage (MB)} \\
			\hline
			Samsung SM-G991U1 & Snapdragon 888 (API 35) & 7.4 GB & 405.5 & 2.5 & 22.4 \\
			\hline
			Samsung SM-A075F & MediaTek Helio G99 (API 36) & 3.6 GB & 480.8 & 2.1 & 14.3 \\
			\hline
			TECNO KM5 & MediaTek Helio G85 (API 35) & 3.7 GB & 724.8 & 1.4 & 16.6 \\
			\hline
			Infinix X6517 & UNISOC SC9863A (API 31) & 3.9 GB & 1192.7 & 0.8 & 22.3 \\
			\hline
			Itel A663L & UNISOC SC9863A (API 33) & 1.9 GB & 1759.4 & 0.6 & 15.6 \\
			\hline
		\end{tabularx}
	\end{table}
	
	\subsection{Error Analysis, Class Stratification, and Confusion Matrix}
\begin{figure}[htbp]
		\centering
		\includegraphics[width=0.65\linewidth]{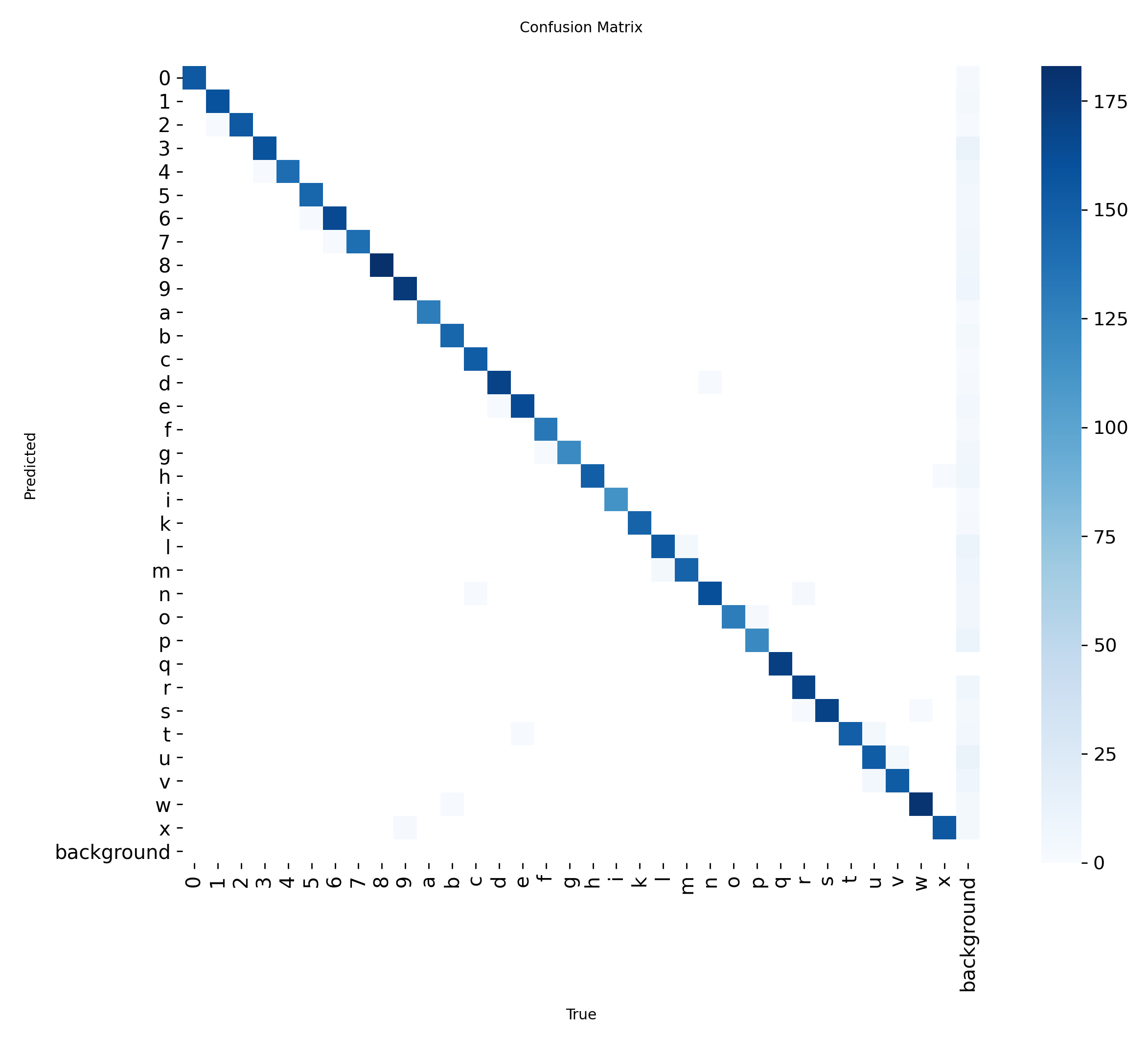}
		\caption{Absolute confusion matrix detailing raw prediction instances across the 33 static alphanumeric sign classes \cite{adeleke2026nsl}.}
		\label{fig:cm_abs}
	\end{figure}

\begin{figure}[htbp]
		\centering
		\includegraphics[width=0.65\linewidth]{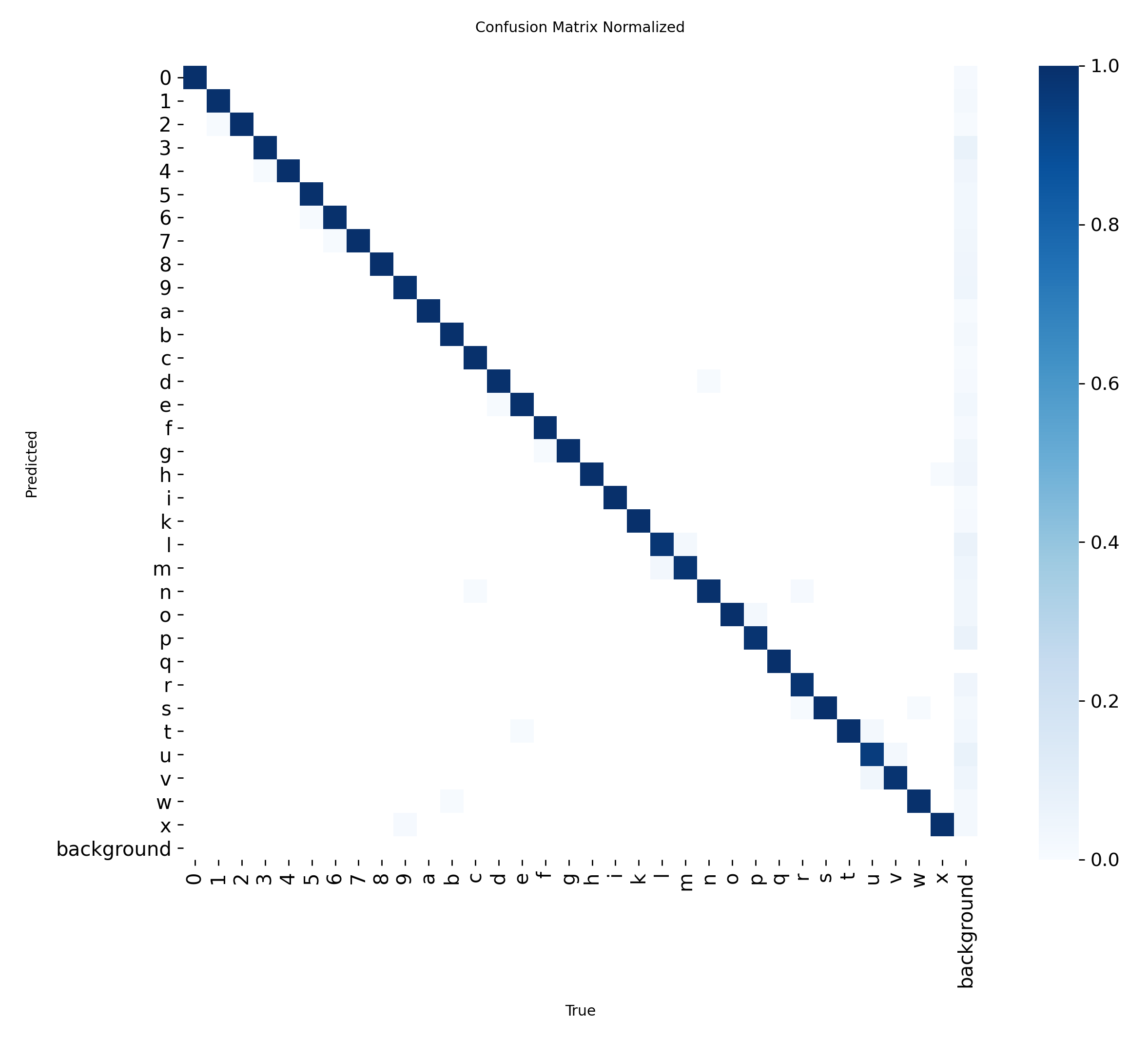}
		\caption{Normalized confusion matrix depicting class-wise accuracy ratios across the 33 categories \cite{adeleke2026nsl}.}
		\label{fig:cm_norm}
	\end{figure}

	The 33 target classes maintain strict diagonal accuracy on the confusion matrix \cite{adeleke2026nsl}. The absolute confusion matrix and the normalized confusion matrix are displayed in Figure~\ref{fig:cm_abs} and Figure~\ref{fig:cm_norm}, respectively \cite{adeleke2026nsl}. The network restricts misclassifications to physically similar hand shapes \cite{pathan2023sign}. The letters 'm' and 'n' cause the only major classification overlap \cite{adeleke2026nsl}. Stratified analysis isolates recall degradation to gestures measuring high on the Spatial-Density Occlusion Index \cite{baihan2024sign}. These specific classes experience a measured 1.2\% error rate \cite{adeleke2026nsl}. Physical self-occlusion from tightly overlapping fingers forces this exact misclassification penalty \cite{pathan2023sign}.
	
	\begin{figure}[htbp]
		\centering
		\includegraphics[width=0.6\linewidth]{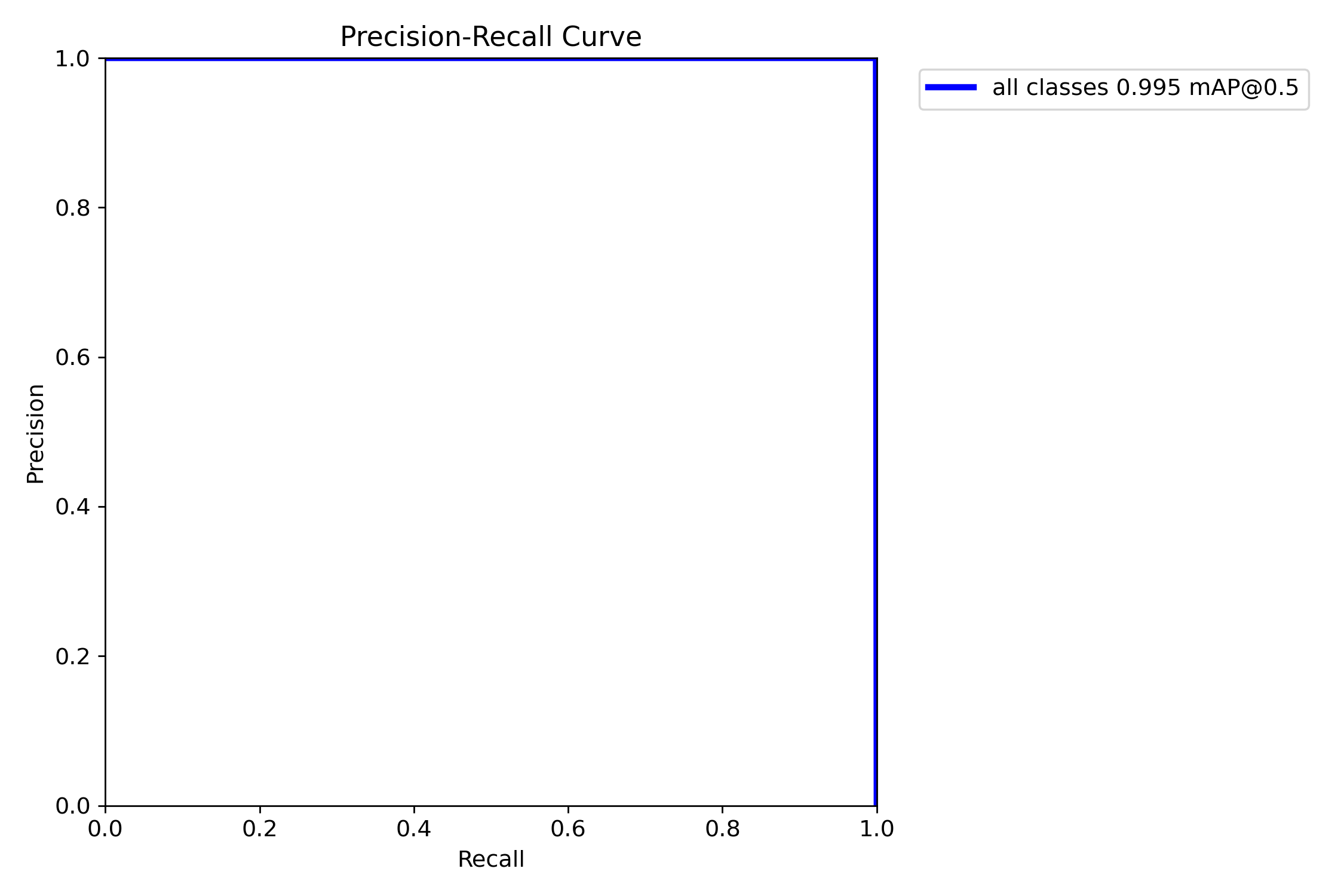}
		\caption{Precision-Recall curve demonstrating overall class mAP@0.5 performance \cite{adeleke2026nsl}.}
		\label{fig:pr_curve}
	\end{figure}
	
	\subsection{Discussion and Practical Implications}
	Single-stage detection networks execute inference directly on mobile hardware \cite{redmon2016you, terven2023comprehensive}. This localized computing topology completely eliminates network transmission latency \cite{akinponnle2021cloud}. Strict bounding boxes isolate the target hand geometries \cite{pathan2023sign}. This spatial constraint shields the primary detector from background environmental interference \cite{baihan2024sign}. Float16 half-precision quantization circumvents rigid physical memory limits \cite{khanam2024yolov11}. Compressing the model drops memory demands to levels easily processed by budget entry-level chips \cite{alishzade2025sign}.
	
	We trained the network architecture entirely on Nigerian Sign Language alphanumeric fingerspelling data \cite{adeleke2026nsl}. The core structural feature extractors map directly to cross-regional vocabularies \cite{gueuwou2025afrisign}. Adapting the model for American or Indian Sign Language variants requires only targeted fine-tuning of the terminal classification layers \cite{navin2025bilingual}. The current framework parses static hand postures \cite{adeleke2026nsl}. Future network builds will introduce temporal tracking algorithms to process fluid dynamic signing motions \cite{bhuiyan2025enhancing}.
	
	\section{Conclusion}
	Float16 compression reduces the YOLO11n detector to a 5.1 MB storage footprint \cite{adeleke2026nsl, khanam2024yolov11}. The quantized architecture maintains a 0.9942 mAP50 evaluation across 33 alphanumeric sign language classes \cite{adeleke2026nsl}. This reduced mass permits installation on heavily constrained mobile hardware \cite{alishzade2025sign}. Telemetry logs from five physical Android smartphones confirmed isolated offline processing \cite{adeleke2026nsl}. The system requires very little active memory \cite{khanam2024yolov11}. All visual frame processing occurs strictly on the local processor \cite{alishzade2025sign}. This edge compute model removes any reliance on cloud data centers \cite{akinponnle2021cloud}. It creates a deployment strategy for assistive tools operating outside reliable broadband networks \cite{olatunji2025perception}.
	
	\section*{Acknowledgments}
	We acknowledge the sign language users who contributed the foundational visual data \cite{adeleke2026nsl}. Their participation anchored the experimental design \cite{adeleke2026nsl}. 
	
	\section*{Declarations}
	\begin{itemize}
		\item \textbf{Funding:} No external funding was received for this study.
		\item \textbf{Conflict of Interest:} The authors declare no conflicts of interest.
		\item \textbf{Analytical / Ethical Approval:} Ethical approval was granted by Wesley University Directorate of Research under protocol number WU/IRB/2025/014 \cite{adeleke2026nsl}.
		\item \textbf{Informed Consent:} Written informed consent was obtained from all participants \cite{adeleke2026nsl}.
		\item \textbf{Clinical Trial Registration:} Not applicable.
		\item \textbf{Consent to Publish Declaration:} Not applicable.
		\item \textbf{Data and Code Availability:} The dataset is available on \href{https://www.kaggle.com/datasets/adelekeadeniyan/nsl-alphanumeric-dataset}{Kaggle} and Zenodo \cite{adeleke2026nsl}. Implementation code is hosted on \href{https://github.com/adelekeadeniyan/NSL-Alphanumeric-Real-time-Sign-language-detection-using-YOLO}{GitHub} \cite{adeleke2026nsl}.
	\end{itemize}
	
	\clearpage
	\bibliographystyle{plain}
	\bibliography{reference}
	
\end{document}